\documentclass[sn-mathphys,Numbered]{sn-jnl}

\usepackage{graphicx}%
\usepackage{multirow}%
\usepackage{amsmath,amssymb,amsfonts}%
\usepackage{amsthm}%
\usepackage{mathrsfs}%
\usepackage[title]{appendix}%
\usepackage{xcolor}%
\usepackage{textcomp}%
\usepackage{manyfoot}%
\usepackage{booktabs}%
\usepackage{algorithm}%
\usepackage{algorithmicx}%
\usepackage{algpseudocode}%
\usepackage{listings}%
\usepackage{geometry}
\theoremstyle{thmstyleone}%
\theoremstyle{thmstyletwo}%

\theoremstyle{thmstylethree}%

\begin{document}

\title[Article Title]{\textbf{Cardiac Arrhythmia Detection using Artificial Neural Network 
}}


\author*[3]{\fnm{Kishore Anand} \sur{K}}\email{kishoreanand.k2019@vitstudent.ac.in}

\author*[4]{\fnm{Sreevatsan} \sur{B}}\email{sreevatsan.b2019@vitstudent.ac.in}

\author*[5]{\fnm{Vishal Kumar} \sur{A}}\email{vishalkumar.a2019@vitstudent.ac.in}
\author*[1]{\fnm{Sangeetha} \sur{R. G.}}\email{sangeetha.rg@vit.ac.in}
\affil[1,1,1,1]{\orgdiv{Electronics and Communication Engineering Department}, \orgname{Vellore Institute of Technology, Chennai}, \orgaddress{\street{Vandalur - Kelambakkam Road}, \city{Chennai}, \postcode{600048}, \state{Tamil Nadu}, \country{India}}}


\abstract{The prime purpose of this project is to develop a portable cardiac abnormality monitoring device which can drastically improvise the quality of the monitoring and the overall safety of the device. While a generic, low cost, wearable battery powered device for such applications may not yield sufficient performance, such devices combined with the capabilities of Artificial Neural Network algorithms can however, prove to be as competent as high end flexible and wearable monitoring devices fabricated using advanced manufacturing technologies. This paper evaluates the feasibility of the Levenberg-Marquardt ANN algorithm for use in any generic low power wearable devices implemented either as a pure real-time embedded system or as an IoT device capable of uploading the monitored readings to the cloud.}




\maketitle

\section{Introduction}\label{sec1}

The pattern of cardiac rhythms has the implying that whether the heart beats typically or not. A healthy heart can effectively circulate blood throughout the body with a normal heartbeat. Heart surrenders alludes to the shortcomings and downfalls in the heart and its vasculature. These imperfections incorporate coronary illnesses and myocardial areas of localized necrosis. Heart abandons are for the most part liable for the aggravations in the normal pulsates of the heart, which is called arrhythmias \cite{bib1}.

The World Health Organization approximates that about 12 million deaths occur worldwide every year subsequently of Coronary Heart Diseases (CHD). Several aspects such as diabetes, HDL \& LDL cholesterol, smoking, obesity, family history makes a person more susceptible to CHDs. In these scenarios, timely prediction and diagnosis of cardiac abnormalities proves to be absolutely necessary for people who are prone to such conditions.

In the region of Nagpur, India\cite{bib9}, the study authors conducted a screening camp for artial fibrillation (AF) and collected essential data including blood pressure, height, weight, diabetes check and the ECG of the residents. Although AF is a public health issue of utmost importance throughout the world\cite{bib10}, this study also notes that a single study\cite{bib11} conducted in a remote Himalayan village residents is the only large-scale population-based study of artial fibrillation in India prevailing until their paper.

Artificial Neural Network is a critical tool for evaluation and prediction. ANN based consumer solutions in the medical industry plays an essential role in improvising the performance and credibility of medical management of general population. ANNs have the potential to sense for various critical conditions, in this case, various CHDs and arrhythmias. The surplus availability of the corresponding conditions' monitored readings (data sets) in electronic form means that the necessary resources to train an ANN model can be deployed with a significant reduction in both development time and time to market.

Since the equated model is trained to record the patterns and rhythms of the ECG wave-forms, ECG wave-forms have been obtained and processed through ANNs for the diagnosis of cardiovascular diseases. Disease detection is based on the P Q R S intervals. Despite the fact that there were a number of other obstacles in the way, the final accuracy was only 71\%, which is why multi-layer perceptron (MLP) backtracking began to be used in disease detection shortly after its introduction. With the assistance of ANNs, fully automated procedures were compared to the clinical interpretation.

The overview of this paper is assembled as follows. Section 2 lists the essential literature surveys. Section 3 gives an insight into the implementation methodology. Section 4 describes the Levenberg-Marquardt ANN model. Section 5 evaluates the final results and the general efficacy of the said ANN model in terms of accuracy and feasibility. The final conclusion of this paper is mentioned in Section 6.

\section{Literature Survey}\label{sec2}

In this study \cite{bib2}, 500 pregnant women's NI-FECG and echocardiography recordings were gathered by the authors. The fetal echocardiography reference diagnostic and the NI-FECG-based diagnosis were compared after the cardiologists examined the extracted NI-FECG. This study comes to the conclusion that while it may be possible to detect the fetal arrhythmias with the of this NI-FECG technique, it is crucial to emphasise the improvements in the models that are used to reconstruct the P-wave.

In another study \cite{bib3}, the authors have given an insight into the significance of using an Apple Watch to clinically monitor the heart beat rates.

Yet another study \cite{bib8} discusses about the Intrapartum fetal cardiac arrhythmias. Using a direct fetal ECG, the authors of this paper gathered 15 incidences of fetal cardiac arrhythmia. According to the study, there are 12.4 arrhythmias for every 1000 observed newborns. The report also mentions that any arrhythmia was associated with the varied decelerations.

This particular study \cite{bib4}, the authors have evaluated the transpiring aspects of electronic wearable technologies in the detection of arrhythmia. This paper evaluates the performance of photoplethysmography (PPG) based wearable devices with and without the aid of machine-learning algorithms with the authors concluding that the latter proved to give a more reliable performance.

The above studies were essential for this paper.

\section{Implementation Methodology}\label{sec6}
This paper describes our implementation of a real-time embedded battery powered working model. The all-inclusive block diagram is demonstrated in the following figure: \\
\begin{center}
    \includegraphics[scale=0.5]{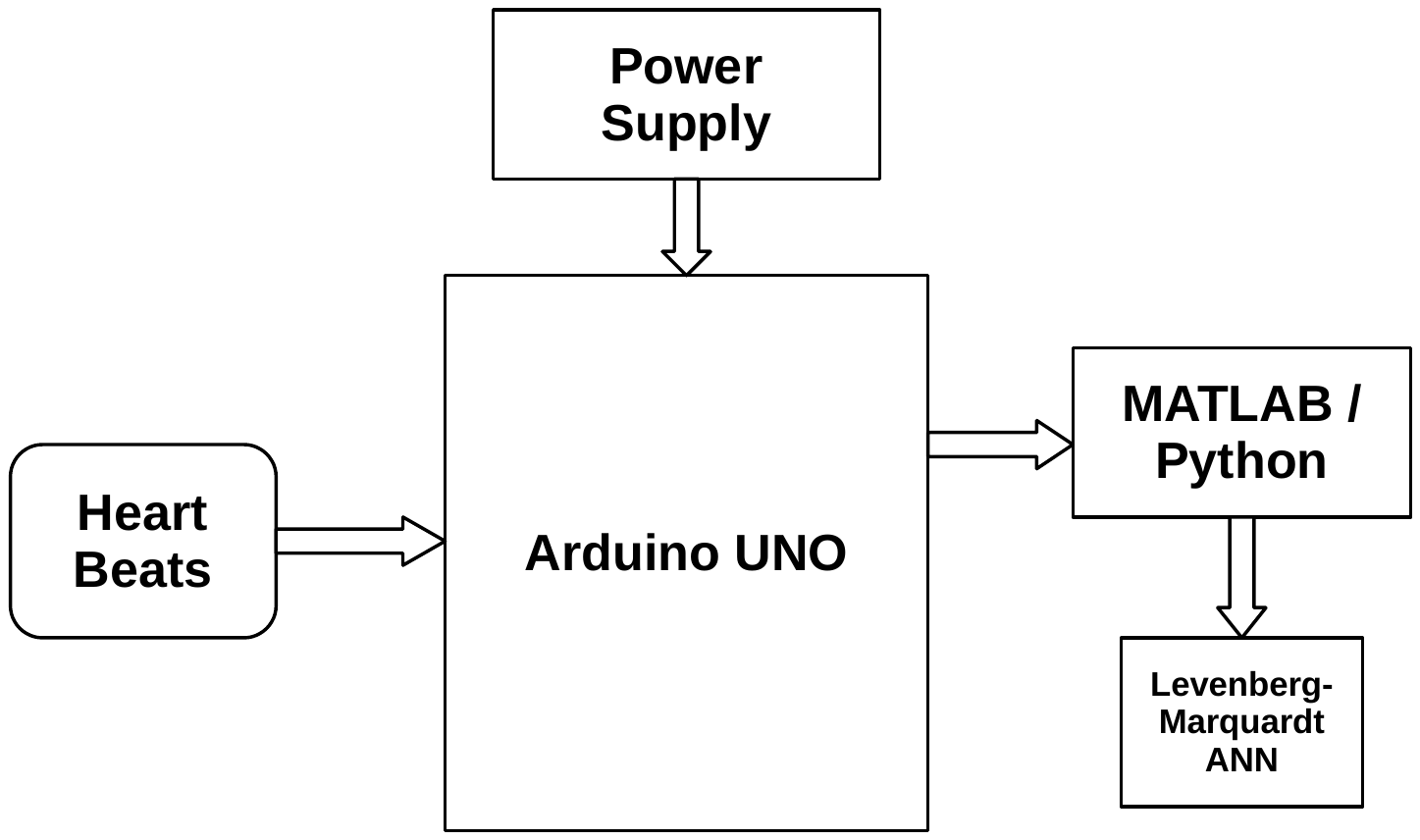}
\end{center}
The working model comprises of an Arduino UNO development board (with ATmega328P as its 8-bit microcontroller), a generic heart sensor - for monitoring the heart beats in real-time and a power source. The monitored readings will be stored in a Microsoft Excel spreadsheet and this file is used as a database for the ANN.

In the context of this paper, the Jupyter Notebook Python tool has been used to implement the Levenberg-Marquardt algorithm to predict the prevalence of any underlying cardiac abnormalities along with the predicted 'death rates'.

\section{Levenberg-Marquardt ANN Model}\label{sec3}

The Levenberg-Marquardt (LM), an advanced non-linear optimization algorithm \cite{bib5} is one of the fastest algorithm available for MLPs. The principal application of Levenberg-Marquardt algorithm lies on solving the issue of fitting least squares curve which is also sketched for minimizing the sum of square error functions. The LM algorithm optimises the variables of the curve of the mentioned model $f(x,\beta)$ in a known realistic data collection of variables, $(x_i, y_i)$, whereupon one can achieve infinitesimally the product of the squared values of the deviations $S(\beta)$ becomes.

\begin{equation}
    S(\beta)=\sum_{i=1}^{a}[y_i-f(x_i,\beta)]^2
\end{equation}

The parameter vector, $\beta$, is changed in each weight change step by a fresh approximation, $(\delta + \beta)$. The linear equivalent of the functions $f(x_i, \delta + \beta)$ is used to determine $\delta$ \cite{bib6}.

\begin{equation}
    f(x_i, \delta + \beta) \thickapprox J_i \delta + f(x_i, \beta)
\end{equation}

Wherein the variables: $J_i=\frac{f(x_i, \beta)\delta }{\beta\delta}$ is the descent of the function $f$ with respect to the parameter vector $\beta$. \newline

Regarding, we select the derivative of $S$. To obtain the smallest values of the squares added together, $S(\beta)$, use this $\delta$ value and set it to zero. The resulting equation is:

\begin{equation}
    (\lambda I + J^TJ)\delta = J^T[y - f(\beta)]
\end{equation}

\begin{equation}
    J = \begin{bmatrix}
    \frac{\delta y_1}{\delta x_1} & ... & \frac{\delta y_1}{\delta x_b}\\
    : & : & :\\
    \frac{\delta y_a}{\delta x_1} & ... & \frac{\delta y_m}{\delta x_b}
    \end{bmatrix}
\end{equation}\\

Where $J$ is the Matrix of Jacobean, $I$ is the Matrix of Identity, $\delta$ as the incrementing factor added to $\beta$ and $\lambda$ is the damping factor.

A smaller value can be used if the fluctuation of S is quick, putting the process closer to the Gauss-Newton technique. One single step nearer to the gradient descent approach can be achieved for modest changes in S by increasing $\lambda$ \cite{bib7}. The diagonal matrix made up of the $J^TJ$ diagonal elements replaces the identity matrix $(I)$ in the Levenberg-Marquardt method.

With the use of the LM approach, gradient components can be scaled in accordance with curvature to allow for bigger swaying in the direction of minor gradients and prevent the slow convergence on these directions.

\section{Algorithm Evaluation and Results}\label{sec4}

\subsection{Input Data Analysis}
The database used to test the performance and working of the Levenberg-Marquardt ANN model contains hundreds of patient's clinical record. For the reason of simplicity and clarity, only 5 of it is mention in the following tabulation:\\

\begin{center}
    \begin{tabular}{|| c c c c c c c c c ||}
    \hline
    Age & Anaemia & Diabetes & High BP & Platelets & Sex & Smoking & Time & Death Event\\
    \hline\hline
    75 & 0 & 0 & 1 & 265000 & 1 & 0 & 4 & 1 \\
    \hline

    55 & 0 & 0 & 0 & 263358.03 & 1 & 0 & 6 & 1 \\
    \hline

    65 & 0 & 0 & 0 & 162000 & 1 & 1 & 7 & 1 \\
    \hline

    50 & 1 & 0 & 0 & 210000 & 1 & 0 & 7 & 1 \\
    \hline

    65 & 1 & 1 & 0 & 327000 & 0 & 0 & 8 & 1 \\
    \hline
    \end{tabular}
\end{center}

\subsection{Output Results}
The complete database have been used to train and validate the Levenberg-Marquardt model. The following figure shows the data representation in Jupyter Notebook Python:\\
\begin{center}
\includegraphics[scale=0.65]{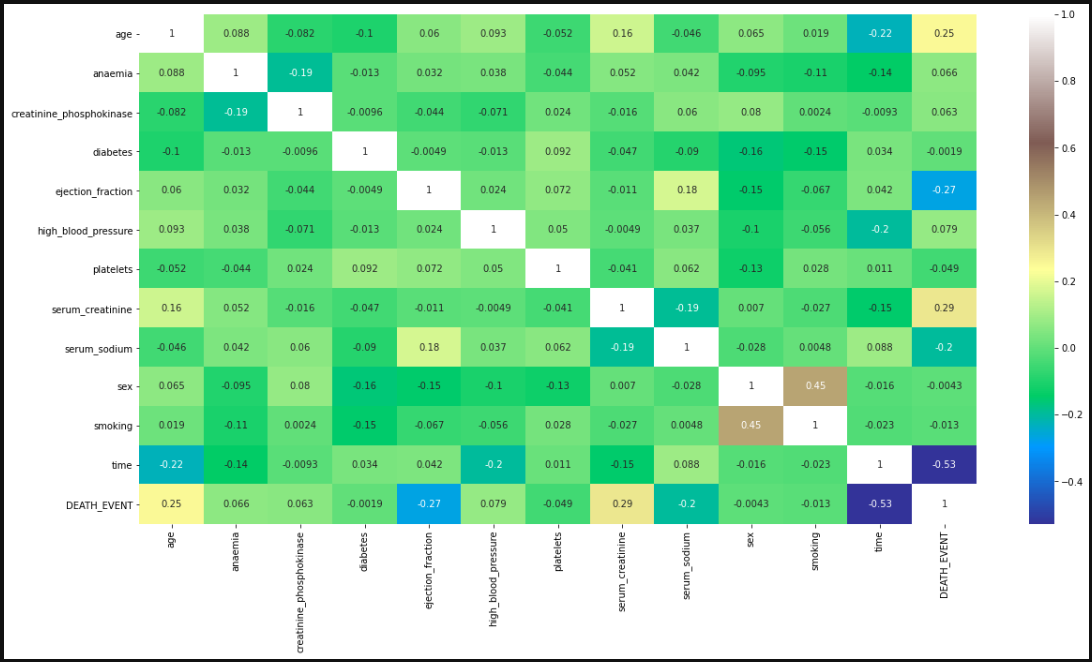}\\ \hfill \break
\end{center}
The following figures represent the accuracy of prediction of cardiac abnormality:\\
\begin{center}
\includegraphics[scale=1.5]{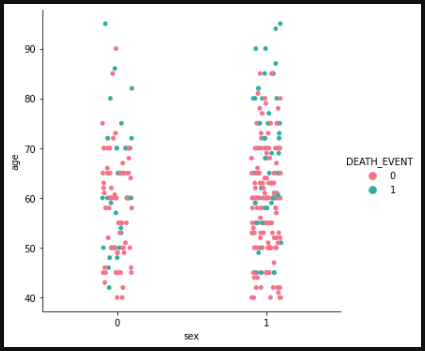}\\
\includegraphics[scale=1.50]{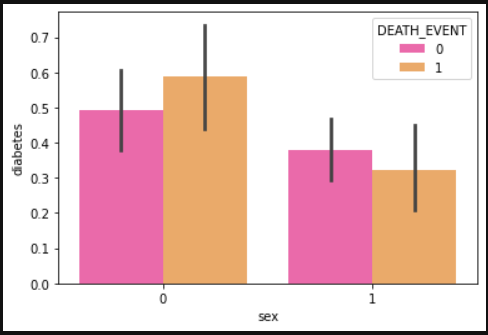}\\ \hfill \break
\hfill \break
\includegraphics[scale=1.45]{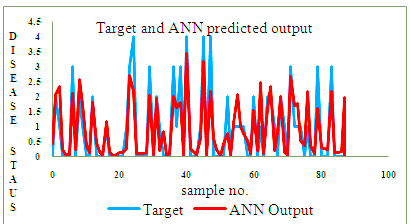}\\
\hfill \break
\hfill \break
\hfill \break
\hfill \break
\hfill \break
\end{center}
The figure for confusion matrix of the trained model is as follows:\\
\begin{center}
\includegraphics[scale=1.25]{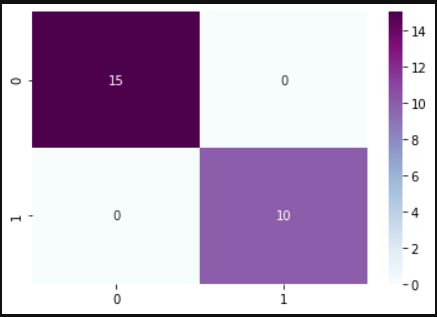}\\
\hfill \break
\end{center}

\section{Conclusion \& Future Work}\label{sec5}

Taking everything into account, when the ANN was prepared, approved and tried in the wake of upgrading the information boundaries, the generally speaking prescient precision acquired was 93.034

The selected database has been used to confirm the accuracy of the system's generated results. As a result, the proposed ANN model can successfully identify individuals at risk for coronary heart disease and predicts future risks for patients.\\

\bibliography{sn-bibliography}

\end{document}